%% file: main.tex
\documentclass[10pt,twocolumn,letterpaper]{article}

\usepackage[pagenumbers]{cvpr} % To force page numbers, e.g. for an arXiv version
\usepackage[T1]{fontenc}


\definecolor{cvprblue}{rgb}{0.21,0.49,0.74}
\usepackage[pagebackref,breaklinks,colorlinks,allcolors=cvprblue]{hyperref}

\def\paperID{51} % *** Enter the Paper ID here
\def\confName{EarthVision}
\def\confYear{2025}

\title{Mapping biodiversity at very-high resolution in Europe}

\author{%
  {César Leblanc}$^1$, {Lukas Picek}$^1$, {Benjamin Deneu}$^2$, {Pierre Bonnet}$^3$, \\ {Maximilien Servajean}$^4$, {Rémi Palard}$^3$, and {Alexis Joly}$^1$\\
  {\small$^1$ INRIA, $^2$ WSL, $^3$ CIRAD, and $^4$ LIRMM}
}

\begin{document}
\twocolumn[{
\renewcommand\twocolumn[1][]{#1}
\maketitle
\centering
\captionsetup{type=figure}
\vspace{-0.5cm}
\includegraphics[width=\textwidth]{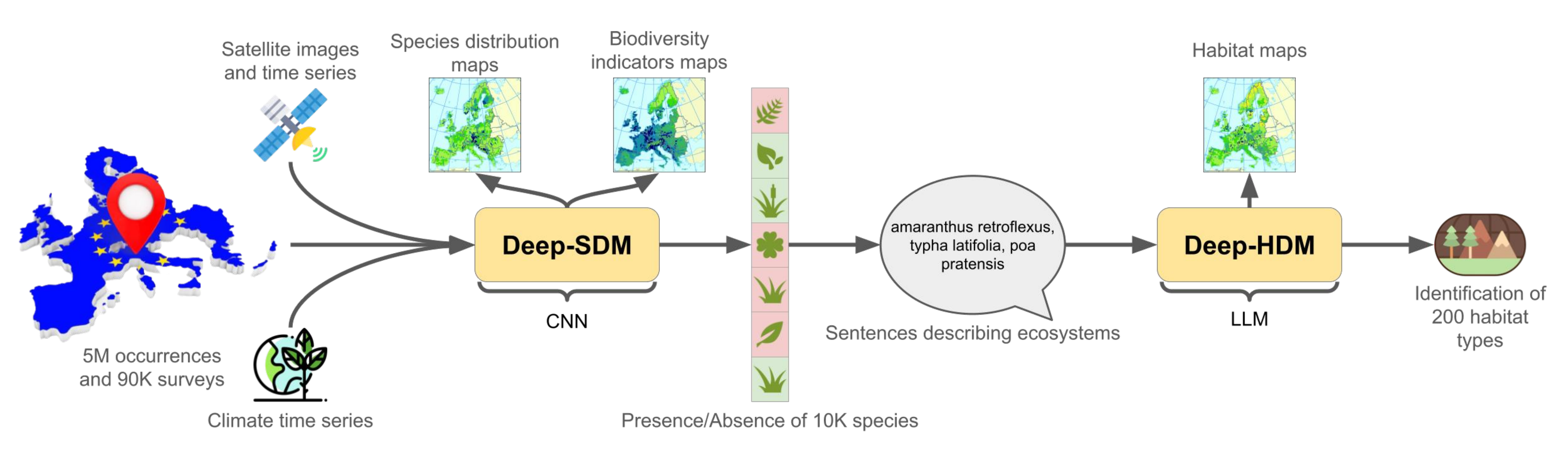}
% \captionof{figure}{Graphical abstract of the cascading approach leveraged in this paper.}
% \label{fig:graphical abstract}
\vspace{0.1cm}
\hypertarget{fig:graphical abstract}{}
}]
\input{sec/0_abstract}    
\input{sec/1_introduction}
\input{sec/1.2_relatedwork}
\input{sec/2_methodology}

\input{sec/3_results}

\input{sec/4_conclusion}
\input{sec/X_acknowledgments}
{
    \small
    \bibliographystyle{ieeenat_fullname}
    \bibliography{main}
}

\end{document}

%% file: sec/0_abstract.tex
\begin{abstract}
This paper describes a cascading multimodal pipeline for high-resolution biodiversity mapping across Europe, integrating species distribution modeling, biodiversity indicators, and habitat classification. The proposed pipeline first predicts species compositions using a \href{https://github.com/plantnet/GeoPlant}{deep-SDM}, a multimodal model trained on remote sensing, climate time series, and species occurrence data at 50$\times$50m resolution. These predictions are then used to generate biodiversity indicator maps and classify habitats with \href{https://github.com/cesar-leblanc/PlantBERT}{Pl@ntBERT}, a transformer-based LLM designed for species-to-habitat mapping. With this approach, continental-scale species distribution maps, biodiversity indicator maps, and habitat maps are produced, providing fine-grained ecological insights. Unlike traditional methods, this framework enables joint modeling of interspecies dependencies, bias-aware training with heterogeneous presence-absence data, and large-scale inference from multi-source remote sensing inputs. 
\end{abstract}

%% file: sec/1_introduction.tex
\section{Introduction}
\label{sec:intro}

% Downstream task. Why is it important. Why we need to do it.
Mapping biodiversity at high spatial resolution is essential for monitoring ecosystem health, assessing species distributions, and guiding conservation policies \cite{pettorelli2016framing,jetz2012integrating,guisan2013predicting}. Effective biodiversity mapping enables the early detection of habitat loss, ecosystem degradation, and climate-induced changes in species ranges, providing crucial information for ecological research and decision-making \cite{newbold2015global,bellard2012impacts,parmesan2003globally}. 

However, generating such maps at a continental scale with fine spatial detail remains a significant challenge due to the limited availability of structured \textit{in situ} data (i.e., data collected directly in the field), spatial biases in species observations, and the complex relationships between species and environmental factors ~\cite{meyer2016multidimensional,deneu2021convolutional,elith2009species}.

% How is it usually or how it could be solved.
A standard approach to tackle these challenges is integrating publicly available species occurrence data, ecological surveys, and remote sensing datasets. Citizen science platforms such as \href{https://www.gbif.org/}{GBIF}, \href{https://plantnet.org/}{Pl@ntNet}, and \href{https://www.inaturalist.org/}{iNaturalist} provide large-scale species presence records ~\cite{bonnet2023synergizing,contini2025seatizen}, while comprehensive biological surveys like \href{https://euroveg.org/eva-database/}{EVA} offer detailed vegetation data, including species composition and habitat characteristics. Additionally, remote sensing data from satellites such as Sentinel and Landsat enable large-scale biodiversity assessments by capturing environmental variables (e.g., precipitation, temperature, and soil) ~\cite{karger2020high} at high spatial and temporal resolutions.
To \textit{convert} these sources to biodiversity maps, Species Distribution Models (SDMs) are widely used ~\cite{guisan2005predicting}. They predict species occurrence by analyzing the relationship between observed records and environmental conditions. Traditional approaches, such as MAXENT \cite{phillips2006maximum} and Random Forest \cite{prasad2006newer} models, rely on statistical correlations but face challenges (e.g., spatial biases, low resolution, and an inability to model species interactions). Recent advances in deep learning-based SDMs (deep-SDMs) overcome these limitations by integrating multi-source data and capturing complex ecological dependencies, resulting in more accurate and scalable biodiversity predictions \cite{deneu2022very}.

% Our approach and contributions.
This work introduces a cascading multimodal pipeline that integrates SDM and Habitat Distribution Modeling (HDM) to generate high-resolution European biodiversity maps. Our approach leverages a deep-SDM, a multimodal model trained on remote sensing (Sentinel-2, Landsat), climate time series, and in situ species observations to predict species compositions at a 50$\times$50m resolution. These predictions form the foundation for computing biodiversity indicator maps, capturing key ecological metrics. Finally, we apply Pl@ntBERT ~\cite{leblanc2024pl}, a transformer-based species-to-habitat classifier, to infer habitat types based on species assemblages, improving habitat mapping beyond traditional remote sensing-based approaches.
Unlike conventional SDMs, which treat species independently and rely on handcrafted environmental features, our deep-SDM models interspecies dependencies, mitigates spatial biases, and enables large-scale inference using heterogeneous presence-absence data. By incorporating HDM, our method extends beyond species distributions to produce detailed habitat maps, providing a more comprehensive view of ecosystem dynamics. This framework offers a scalable and fine-grained solution for biodiversity monitoring, delivering high-resolution species distribution, biodiversity indicators, and habitat maps at a continental scale.

%% file: sec/1.2_relatedwork.tex
\section{Related Work}
\label{sec:related_work}

Accurate biodiversity and habitat mapping have traditionally relied on habitat suitability models \cite{guisan2005predicting} or direct classification from remote sensing data \cite{amani2023three}. However, these methods are often constrained by limited spatial resolution \cite{estopinan2024mapping}, outdated reference datasets, and the inability to model interspecies relationships. Some rare studies combine deep learning, citizen science data, and remote sensing to track plant species changes \cite{gillespie2024deep}. Nevertheless, they are usually geographically restricted to a country.

Since mapping requires models that can predict species distributions and classify habitats. Traditional SDMs estimate species occurrence probabilities using environmental variables, while HDMs focus on habitat classification by analyzing species composition. This section summarizes key facts about SDMs, deep-SDMs, and HDMs, highlighting their strengths, limitations, and relevance to our approach.

\textbf{Species distribution models (SDMs)} predict where species are likely to occur by analyzing relationships between species observations and environmental conditions. Traditional SDMs approaches, i.e., MAXENT and Random Forests, rely on statistical correlations but face limitations, including spatial biases, low resolution, and the inability to model interactions between species \cite{phillips2009sample,brun2024multispecies}. These weaknesses limit their effectiveness, especially when working with large-scale and complex ecosystems.
To overcome these challenges, deep-SDMs integrate remote sensing, climate data, and species occurrences to improve prediction accuracy \cite{deneu2021convolutional,geoplant2024picek,botella2023geolifeclef,dollinger2024sat,sastry2024taxabind}. Unlike traditional SDMs, deep-SDMs can learn complex spatial patterns and ecological relationships using CNNs or transformers. This enables higher-resolution predictions at a large scale, making species distribution modeling more precise and scalable.

\textbf{Habitat distribution models (HDMs)} traditionally rely on expert systems \cite{noble1987role} and machine learning \cite{hastie2009elements}. Expert systems, though widely used \cite{tichy2019grimp}, often overfit, making classification sensitive to minor plot variations, and sometimes require external criteria beyond species composition \cite{de2015comparative}.  
Machine learning models (i.e., NNs) ~\cite{leblanc2024deep}, capture complex species composition patterns \cite{vcerna2005supervised} but treat all species as equally different, failing to model ecological interdependencies \cite{olden2008machine}. While classical approaches are interpretable, they struggle with high-dimensional data.  
Deep learning, particularly transformers \cite{vaswani2017attention}, has shown promise in biology, e.g., protein structure prediction \cite{jumper2021highly}, but remains underexplored in vegetation classification. Their ability to model global dependencies makes them a promising alternative for habitat classification. 

%% file: sec/2_methodology.tex
\section{Methodology}
\label{sec:methodology}

\begin{figure}[h]
  \centering
   \includegraphics[width=\linewidth]{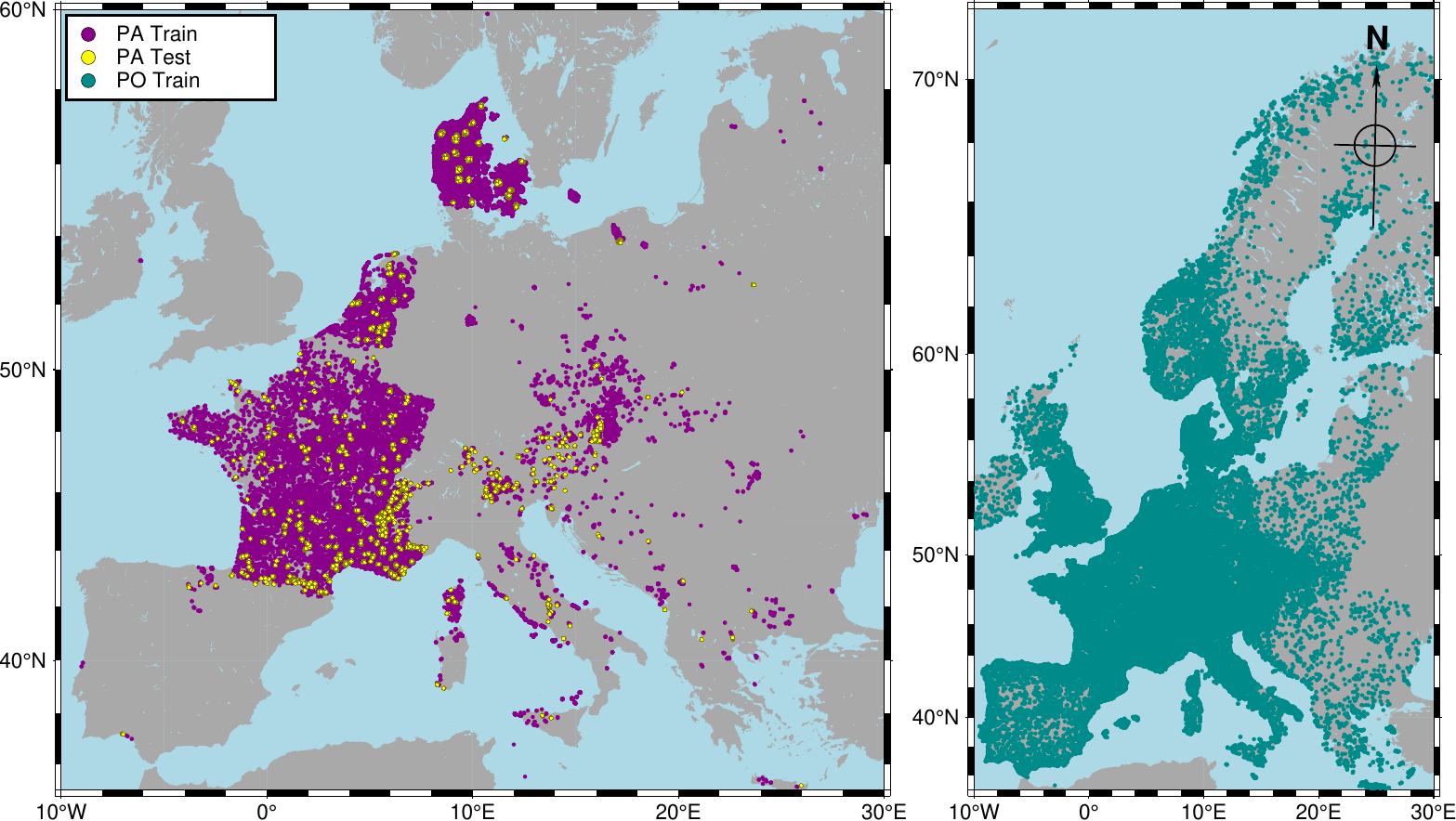}
   \caption{Geo spatial scale of the dataset (from ~\cite{geoplant2024picek}). The 5M PO occurrences (9,709 species) span all of Europe, but the 90K PA surveys (5,016 species) are primarily in France and Denmark.}
   \label{fig:dataset map}
\end{figure}

\paragraph{Dataset.} To construct the maps, we use GeoPlant \cite{geoplant2024picek}, a~new European-scale dataset (see \cref{fig:dataset map}) designed for high-resolution species distribution modeling. GeoPlant covers over 11,000 plant species, i.e., most of the European flora, and is based on 5 million opportunistic Presence-Only (PO) records from GBIF and 90,000 exhaustive Presence-Absence (PA) surveys from the European Vegetation Archive (EVA). Besides, for each plant species observations, Sentinel-2 RGB and NIR satellite images with 10m resolution, a 20-year time series of climatic variables (i.e., precipitation and mean, min, and max month temperature), and satellite time series from the Landsat program (i.e., R, G, B, NIR, and SWIR1+2) are provided. Coordinates were not used as we want to reflect habitat suitability (i.e., learn a relationship between environment and occurrences) \cite{cole2023spatial}.

The provided PO and PA data were aggregated into a 50×50m spatial grid, consistent with the resolution used during inference. This aggregation combines both data types into a single site occupancy dataset, where each grid cell contains a 1 or 0 per species, representing presence or absence. This setup allows using a Binary Cross-Entropy loss function, which is better suited for presence-absence probability estimation than Categorical Cross-Entropy. Additionally, a target group background approach \cite{phillips2009sample} was applied to partially correct sampling bias \cite{barber2022target}. To achieve this, training is restricted to grid cells containing at least one recorded species, ensuring that pseudo-absence points are sampled only from locations where other species have been observed. This method helps compensate for the lack of explicit absence data, improving the ecological relevance of background points.

\paragraph{Species Distribution Modeling.}

Our approach is based on deep multi-modal models, which have been shown to outperform classical SDMs \cite{joly2023overview,joly2024overview,joly2024lifeclef}. The mapping process consists of two main phases: (i) training a deep-SDM using \textit{in situ} observations combined with spatialized environmental and remote sensing data and (ii) inferring the trained model to predict species distributions across Europe.

\begin{figure}[t]
  \centering
   \includegraphics[width=0.95\linewidth]{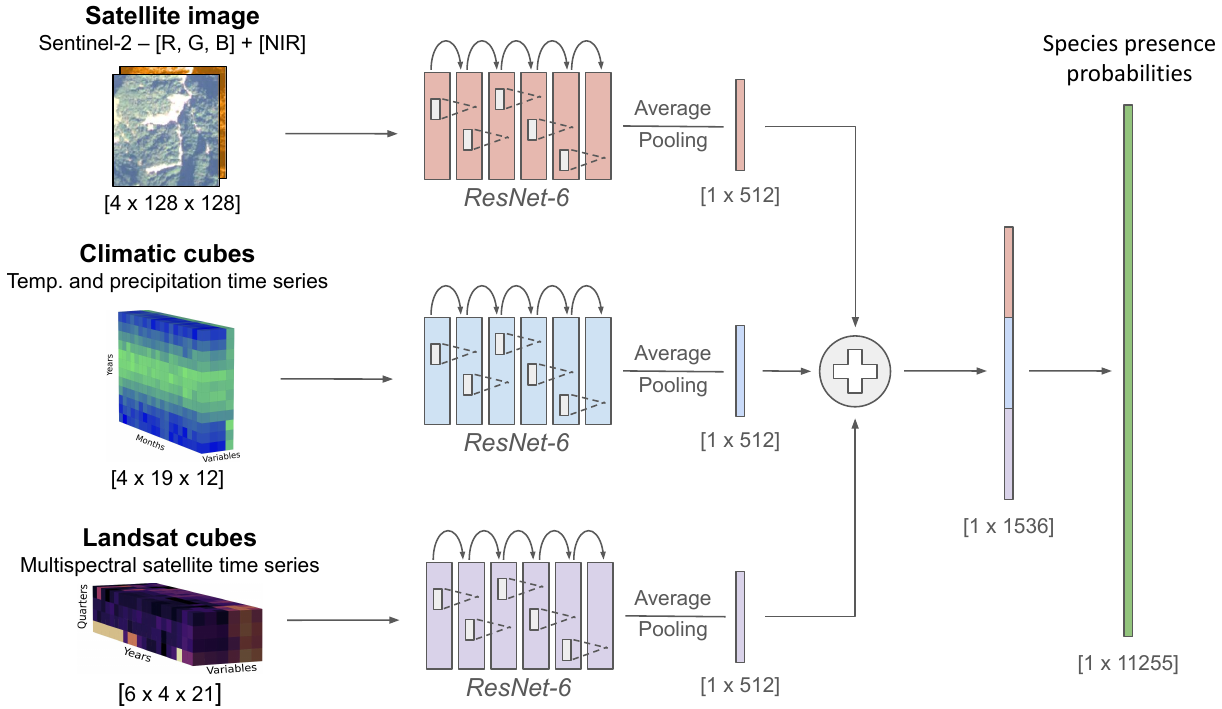}
   \caption{Selected SDM architecture (from ~\cite{geoplant2024picek}). This multimodal ensemble model processes each modality (i.e., satellite images, climatic cubes, and Landsat cubes) through a lightweight 6-layer residual encoder (i.e.,  ResNet-6). The embeddings are then concatenated and passed to a final classification layer.}
   \label{fig:architecture sdm}
\end{figure}

We use a multi-modal ensemble approach (see \cref{fig:architecture sdm}), building on previous work \cite{botella2023overview,picek2024overview,leblanc2022species}, based on a modified ResNet-6 architecture with three separate branches for different input data types: (i) Sentinel-2 RGB+NIR imagery (128$\times$128 patches at 10m resolution), (ii) Climate time series encoded as three-dimensional data cubes (year, month, and variables such as precipitation and temperature), and (iii) Landsat remote sensing time series, structured similarly with spectral bands (R, G, B, NIR, and SWIR1+2).
Each input modality is encoded by a dedicated CNN encoder with six residual blocks, a design choice that improves performance over larger off-the-shelf architectures \cite{geoplant2024picek}. The extracted embeddings are concatenated and passed through a fully connected classifier which computes species presence probabilities using one fully connected layer with a sigmoid activation function.
The model is trained using Stochastic Gradient Descent (SGD) with binary cross-entropy loss. The training code is available on \href{https://github.com/plantnet/GeoPlant}{GeoPlant GitHub}.

\paragraph{Biodiversity Indicator Calculation.}
The biodiversity indicators are extracted from the species assemblages predicted by the SDM at a 50$\times$50m resolution across Europe. These indicators summarize ecological properties such as species richness and the presence of specific taxonomic or functional groups, providing valuable insights into biodiversity patterns.
To derive these assemblages, the species probabilities predicted by the SDM are thresholded using a conformal prediction approach \cite{fontana2023conformal}. This method ensures a low probability of omitting truly present species, even if it results in some false positives. This conservation-focused strategy prioritizes minimizing omission errors and reducing the risk of underestimating species distributions, which is critical for biodiversity assessments.

We define seven biodiversity indicators from the predicted species assemblages to assess ecosystems' conservation status. These indicators capture key ecological and regulatory aspects, e.g.,

\begin{itemize}
    \item \textbf{Species richness}: number of species.
    \item \textbf{EU directive}: number of species from the list provided by the EU Habitat directive.
    \item \textbf{Threatened species}: number of IUCN Red List species.
    \item \textbf{Most threatened}: IUCN threatened species status\footnote{In the case of missing IUCN status, they were inferred using an automated method ~\cite{zizka2022iucnn} also based on neural networks.}.
    \item \textbf{Tree species}: number of species from the lifeform “woody” of the Plan Of the World Online database.
    \item \textbf{Invasive species}: number of species from the CABI list.
    \item \textbf{Specialist species}: number of species estimated to be present with a very low probability of presence elsewhere.
\end{itemize}

Consequently, any indicator $n_S(x)$ relying on a number of present species among $|S|$ species of a particular type, can be modelled as a statistical variable following a Poisson binomial distribution (i.e., a sum of independent Bernoulli trials that are not necessarily identically distributed).
Thus, the mean of $n_S(x)$ can be estimated as
\begin{equation}
    \mu_S(x) = \sum_{i \in S} p(y_i=1|x),
\end{equation}
where $S$ is the set of species of interest (e.g., endangered species) and $x$ is a particular point of the map (i.e., a cell of $50\times50m$). The variance of $n_S(x)$ can be computed as

\begin{equation}
    \sigma_S(x)^2 = \sum_{i \in S} p(y_i=1|x) \cdot (1 - p(y_i=1|x)),
\end{equation}
from which we can derive a confidence interval for each point of the map, e.g., through the 2-sigma rule:
\begin{equation}
    \delta_S(x) = 2\sigma_S(x).
\end{equation}

For an indicator based on \( |S| = 10 \) species with probabilities  
\( p(y_1 = 1 | x) = 0.9 \), \( p(y_2 = 1 | x) = 0.8 \), \( p(y_3 = 1 | x) = 0.1 \), and  
\( p(y_i = 1 | x) = 0 \) for \( i \in [4,10] \), we obtain \( \mu_S(x) = 1.8 \) and  
\( \delta_S(x) = 1.1 \), giving
\begin{equation}
    n_S(x) = 1.8 \pm 1.1.
\end{equation}

Thus, we can build a confidence interval map for almost all indicators (using $\delta_S(x)$ as the value for each point).
Only the IUCN status of the most threatened species does not follow this pattern. For this one, we want to estimate the probability that at least one species of a particular IUCN status is present. If, for instance, we consider the set $S=EN$ of species with status ENDANGERED, the probability that at least one of them is present is equal to
\begin{equation}
    p(n_{EN}(x) > 1 | x) = 1 - \prod_{i \in EN} (1 - p(y_i=1|x)).
\end{equation}

If we have $|EN|=10$ species and $p(y_1=1|x)=0.9$, $p(y_2=1|x)=0.8$, $p(y_3=1|x)=0.1$ and $p(y_i=1|x)=0$ for $i \in [4,10]$, then the probability that at least one ENDANGERED species is present is 98.2\%.

\paragraph{Habitat Identification.} Unlike traditional approaches that train models on satellite imagery labeled with EUNIS habitat types\footnote{The EUNIS habitat classification \cite{moss2008eunis} is a hierarchical system for the categorization of natural and semi-natural habitats in Europe developed to support biodiversity management, conservation, and sustainable use.} \cite{si2025earth}, we infer habitats from the species assemblages predicted by the deep-SDM. Direct habitat classification from remote sensing is limited by the scarcity and outdated nature of labeled datasets, as most available EUNIS labels come from EVA surveys with a mean collection year of 1992. Many labeled sites have undergone significant ecological changes due to land-use transformation and climate change, making direct mapping unreliable. We follow the latest version of the EUNIS classification~\cite{chytry2024floraveg} and focus on levels 1, 2, and 3, the last being the most detailed.

Instead, since the primary value of EVA lies in its plant species assemblage data, we take a different approach: training a supervised model to predict EUNIS habitat types based on species composition. This method is less affected by temporal shifts in habitat labels because species assemblages remain a strong predictor of habitat type, even when direct habitat labels become outdated \cite{leblanc2024deep}. If the deep-SDM accurately predicts species assemblages at a given site, habitat types can be inferred with high confidence.

\begin{figure}[t]
  \centering
   \includegraphics[width=\linewidth]{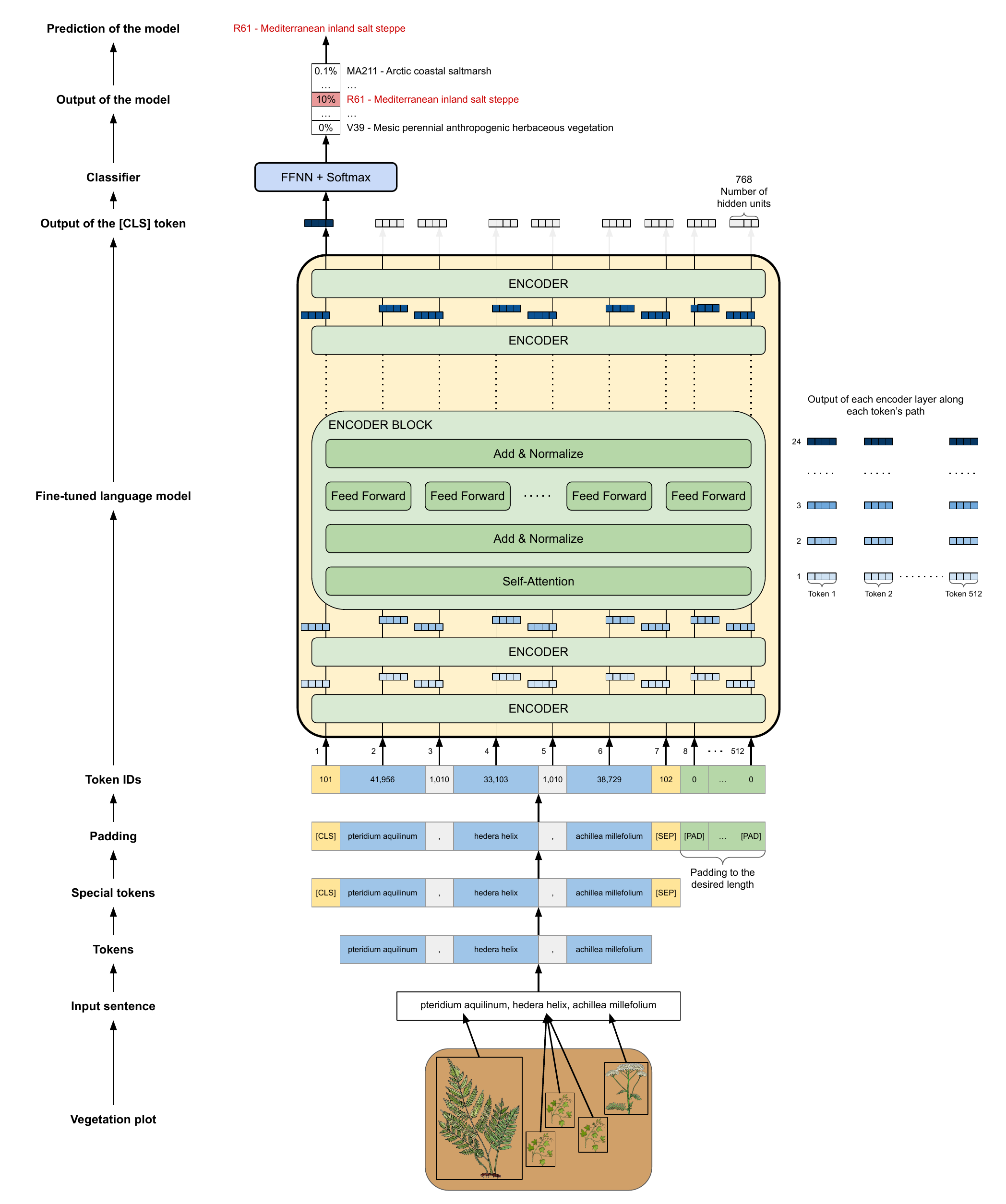}
   \caption{Pl@ntBERT HDM (from ~\cite{leblanc2024pl}) processes the input (list of species predicted by the deep-SDM) through multiple encoder layers, with the [CLS] token representation passed to a classifier to predict the most likely habitat type.}
   \label{fig:architecture hdm}
\end{figure}

To implement this approach, we use Pl@ntBERT, a Python-based framework for training, sharing, and evaluating species-to-habitat classification models. Pl@ntBERT leverages large language models (LLMs), which have demonstrated strong performance in modeling plant species relationships \cite{marcos2024fully}. It is built upon BERT, originally designed for natural language understanding \cite{devlin2019bert}, but adapted to capture latent dependencies between plant species in different ecosystems \cite{morin2009community}.
The model is trained in two stages:

\textit{Species-to-Species prediction}:  Given a predicted species assemblage, Pl@ntBERT learns to recover missing species by training on incomplete species lists. This step refines its understanding of species co-occurrence patterns \cite{gururangan2020don}.

\textit{Species-to-Habitat Classification}: Fine-tuned on species assemblages from the deep-SDM, the model predicts the most probable EUNIS habitat type based on a sorted list of species by estimated spatial coverage (see \cref{fig:architecture hdm}).

Pl@ntBERT provides an efficient and scalable solution for habitat classification, leveraging the predictive power of species assemblages rather than relying on direct but potentially outdated habitat labels. The source code for training and inference is available on \href{https://github.com/cesar-leblanc/PlantBERT}{Pl@ntBERT GitHub}.

%% file: sec/3_results.tex
\begin{figure*}
  \centering
  \begin{subfigure}{\linewidth}
    \centering
    \begin{subfigure}{\linewidth}
      \includegraphics[width=\linewidth]{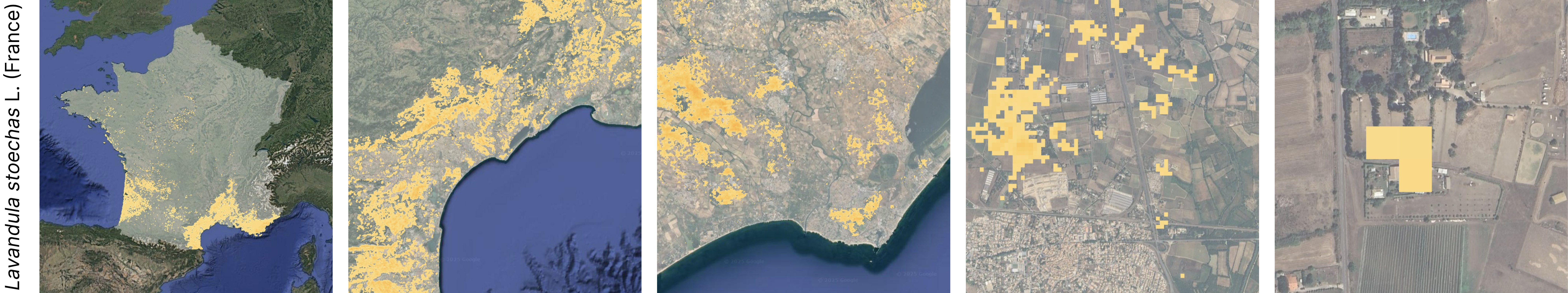}
    \end{subfigure}
    \label{fig:lavandula stoechas}
  \end{subfigure}
  \begin{subfigure}{\linewidth}
    \centering
    \begin{subfigure}{\linewidth}
      \includegraphics[width=\linewidth]{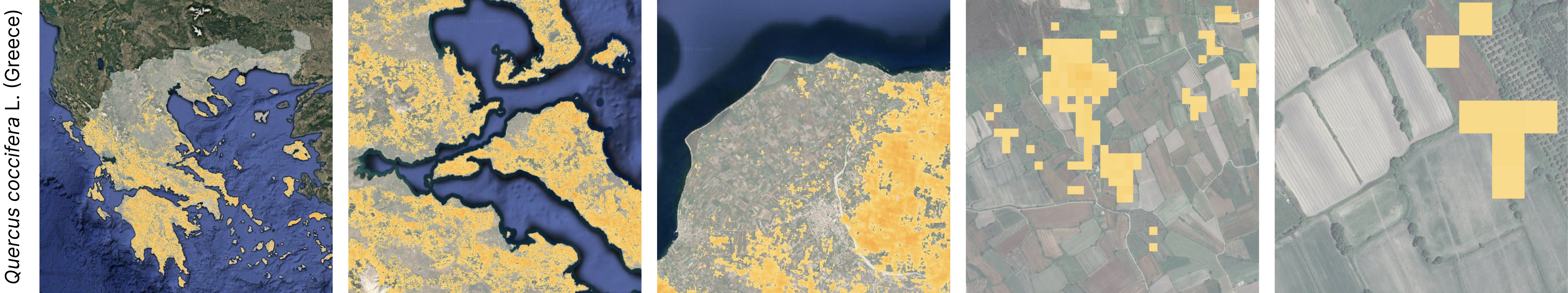}
    \end{subfigure}
    \label{fig:quercus coccifera}
  \end{subfigure}
  \caption{Example species distribution maps for two selected species occurring in France and Greece at different zoom levels. These maps are produced by the deep-SDM all over Europe for over 5,500 plant species at a 50$\times$50m resolution.}
  \label{fig:species distribution maps}
\end{figure*}

\section{Inference Details and Results}
\label{sec:results}

\paragraph{Species Distribution Maps.} The trained SDM was used to generate high-resolution species distribution maps across Europe (see \cref{fig:species distribution maps}). To ensure scalability, the study area was divided into 25$\times$25km meta-tiles, each processed independently. Within each tile, species predictions were made at 50$\times$50m grid, totaling 5.5 billion cells. If a cell’s center fell in water, it was moved to the nearest terrestrial point.

Inference was done for the year 2021, using environmental data averaged between March 21 and December 1, 2021. Unlike in training, where data cubes were extracted based on observation dates (hence capturing seasonal or interannual variability), inference used a fixed reference period. At each inference point, the model predicted presence probabilities for 11,255 species, which were then thresholded to retain only likely present species, significantly reducing storage requirements. The threshold was optimized on the validation set to maximize the F-score.

Maps were generated \textit{just} for 5,558 out of 11,255 species. This does not necessarily indicate species absence but rather that their predicted probability remained below the confidence threshold. On average, a species was predicted in 132.8 million grid cells, covering 332,000 km² (2.4\% of Europe). The most widespread species, \textit{Agrostis capillaris L.}, appeared in 3.23 billion grid cells (58.6\%).

To evaluate model performance, a spatial block hold-out split (10$\times$10km grid) was used to mitigate spatial autocorrelation and assess generalization \cite{roberts2017cross}. This approach was chosen as a realistic test of spatial interpolation based on species occurrence distribution. Each input modality was also evaluated separately in order to demonstrate their own predictive power, with Landsat data resulting in the highest value. See \cref{tab:sdm evaluation} for detailed evaluation.

\begin{figure*}
  \centering
  \begin{subfigure}{\linewidth}
    \centering
    \begin{subfigure}{\linewidth}
      \includegraphics[width=\linewidth]{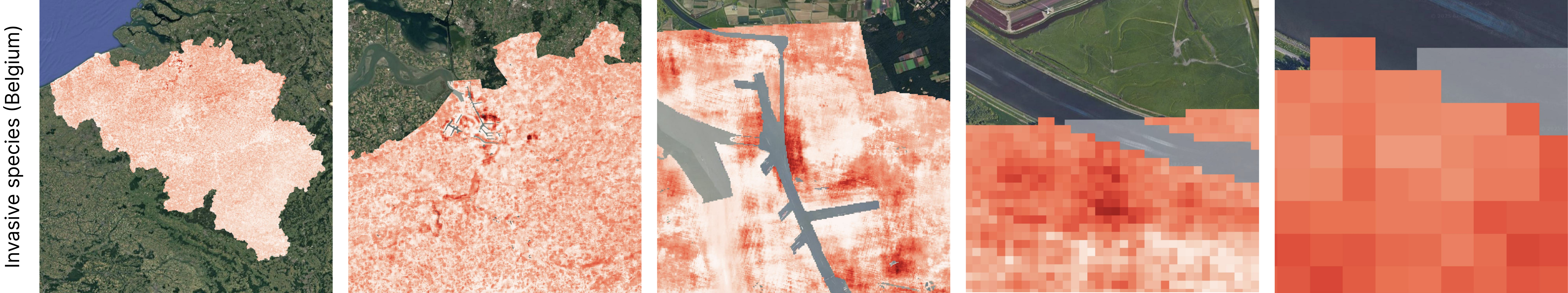}
    \end{subfigure}
    \label{fig:invasive species}
  \end{subfigure}
  \begin{subfigure}{\linewidth}
    \centering
    \begin{subfigure}{\linewidth}
      \includegraphics[width=\linewidth]{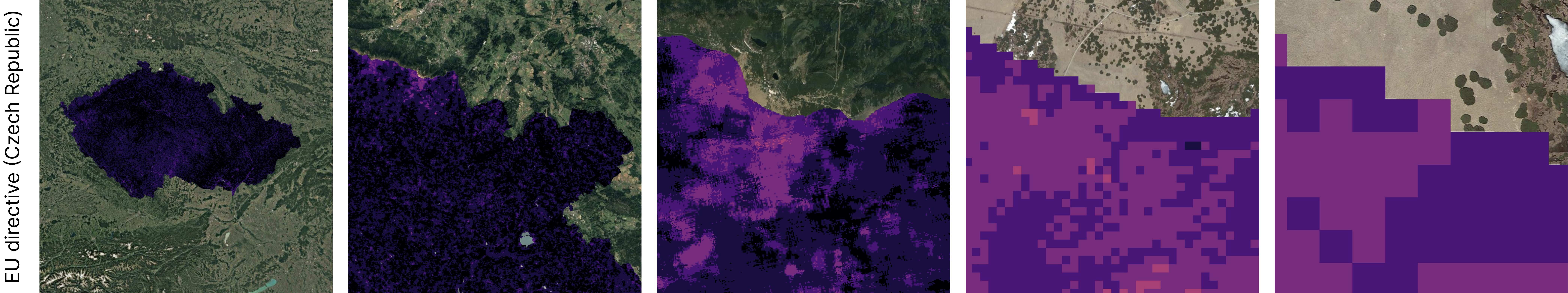}
    \end{subfigure}
    \label{fig:eu directive}
  \end{subfigure}
  \caption{Example biodiversity indicator maps for two selected indicators occurring in Belgium and the Czech Republic at different zoom levels. These maps are produced with the output of the deep-SDM all over Europe for seven biodiversity indicators at a 50$\times$50m resolution.}
  \label{fig:biodiversity indicators maps}
\end{figure*}

The multi-modal model achieves a high AUC score \cite{fawcett2006introduction} of 0.931, indicating strong performance in ranking true presence sites higher than true absence sites. This suggests that the predicted species distribution maps closely align with actual species occurrences. However, the F-score \cite{van1979information}, which requires the model to predict the exact species assemblage for each test plot, is relatively low at 0.338.

A major limitation arises from the scale mismatch between the test vegetation plots and the predicted grid cells. The targeted resolution is 50$\times$50m (2,500m$^2$), whereas test plots average 100m$^2$, meaning they contain significantly fewer species. As a result, many species predicted by the model may be considered false positives (i.e., are over-predicted) at the test plot scale, even if they are present at the full 2,500m$^2$ resolution. \\

\noindent\textbf{Note:} \textit{The full workflow required approximately 30,000 GPU hours on Nvidia A100 GPUs, producing 15TB of data.}

\begin{table}[h]
  \centering
    \caption{Evaluation of the SDM. The multimodal ensemble approach achieves a considerable performance improvement compared to the single modality models in terms of all metrics.}
  \begin{tabular}{@{}lcccc@{}}
    \toprule
    \textbf{Branch} & \textbf{AUC} & \textbf{F-score} & \textbf{Recall@50} & \textbf{Recall@250}\\
    \midrule
    \textit{Sentinel} & 0.898 & 0.258 & 0.524 & 0.848 \\
    \textit{Bio} & 0.891 & 0.273 & 0.544 & 0.872 \\
    \textit{Landsat} & 0.920 & 0.312 & 0.595 & 0.873 \\
    \midrule
    All & \textbf{0.931} & \textbf{0.338} & \textbf{0.639} & \textbf{0.908} \\
    \bottomrule
  \end{tabular}
  \label{tab:sdm evaluation}
\end{table}

A more precise evaluation of recall and precision would require a complete ground-truth dataset at the 2,500m$^2$ scale, which is impossible due to the extreme effort required for manual surveys, or to predict species compositions at a 10$\times$10m resolution, which approximately means multiplying the number of grid cells by 25. Instead, we use Recall@K to evaluate the model’s ability to recover species despite the resolution mismatch. Since the exact number of species in a 50$\times$50m cell is unknown, K=50 and K=250 serve as proxies for low- and high-diversity areas. The model retrieves nearly two-thirds of species for K=50 and over 90\% for K=250, demonstrating strong recall despite the spatial scale limitations. Favoring recall at the expense of false positive ensures species are not missed (key in conservation), even if precision drops.

\begin{figure*}
  \centering
  \begin{subfigure}{\linewidth}
    \centering
    \begin{subfigure}{\linewidth}
      \includegraphics[width=\linewidth]{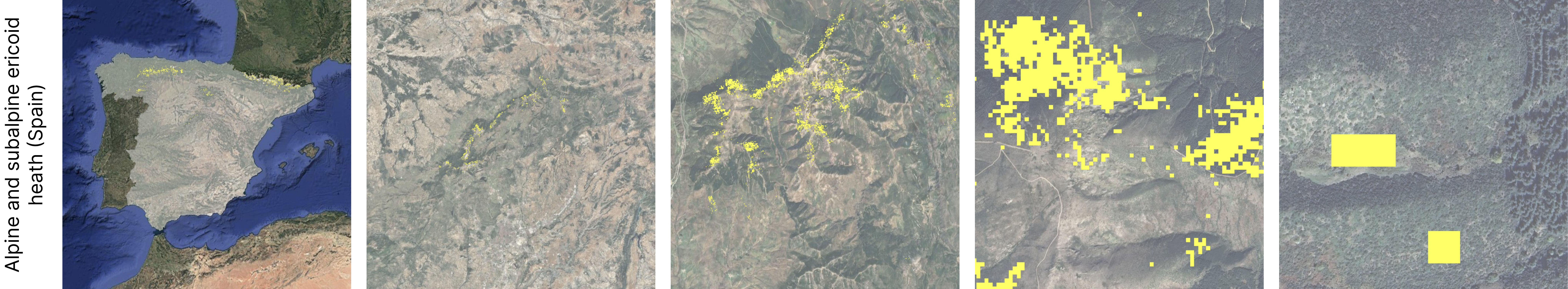}
    \end{subfigure}
    \label{fig:alpine and subalpine ericoid heath}
  \end{subfigure}
  \begin{subfigure}{\linewidth}
    \centering
    \begin{subfigure}{\linewidth}
      \includegraphics[width=\linewidth]{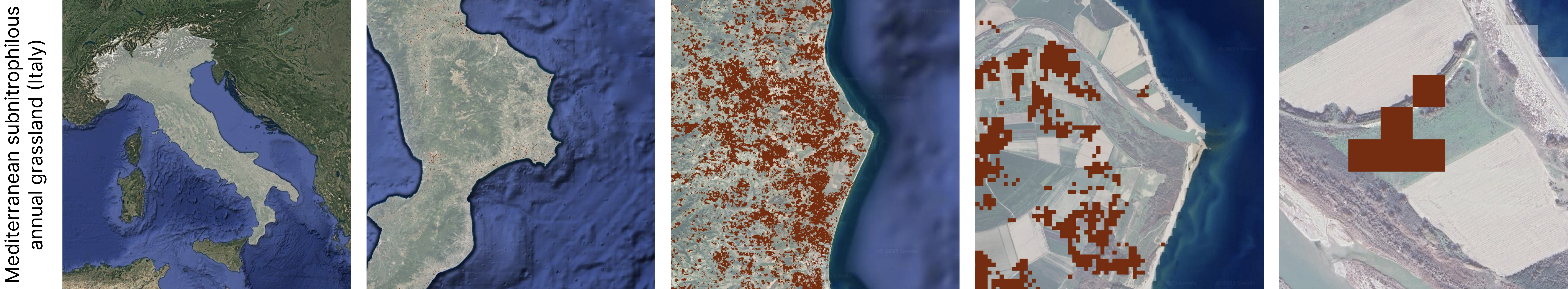}
    \end{subfigure}
    \label{fig:mediterranean subnitrophilous annual grassland}
  \end{subfigure}
  \caption{Example habitat maps for two selected habitats occurring in Spain and Italy at different zoom levels. These maps are produced by Pl@ntBERT with the output of the deep-SDM all over Europe for over 200 habitat types at a 50$\times$50m resolution.}
  \label{fig:habitats maps}
\end{figure*}

\paragraph{Biodiversity Indicators Maps.}
The workflow to create the high-resolution indicator maps at the European scale (see \cref{fig:biodiversity indicators maps}) is closely related to the one used for producing the species distribution maps based on the SDM. The meta-tiles of size 25$\times$25km are processed one by one (in parallel), and within each tile, the indicators are computed for each point of the 50$\times$50m grid based on the species assemblage predicted by the SDM. For most indicators, the two main operations are (i) filtering the species of interest for the targeted indicator and (ii) counting the number of filtered species. This can be implemented very efficiently on a GPU through the use of binary masks and the sum of tensor values. Only the indicator “IUCN status of the most threatened species in the assemblage” requires a slightly different process, but that was efficiently implemented by encoding status as integers and using look-up tables and a max operator. So far, all 7, i.e.,  (i) Species richness, (ii) EU directive, (iii) Threatened species, (iv) Most threatened, (v) Tree species, (vi) Invasive species, and (vii) Specialist species, biodiversity indicator maps have been produced.

\paragraph{Habitat Maps.}
The habitat maps (see \cref{fig:habitats maps}) were inferred following a workflow similar to that used for biodiversity indicators. The study area was divided into 25$\times$25km meta-tiles, which were processed in parallel. Within each tile, the model classified each 50$\times$50m grid cell based on the species probabilities predicted by the SDM. The classifier directly assigns EUNIS Level 3 habitat types, while Levels 1 and 2 are inferred from the hierarchy.

In total, 200 habitat maps were generated at EUNIS Level 3, covering 60.4\% of all habitat types at this level. On average, a habitat was mapped across 27.77 million grid cells, corresponding to 0.49\% of Europe's total area. The most widespread habitat, R22: “Low and medium altitude hay meadow”, was predicted in 681.5 million grid cells, covering 12.27\% of Europe.

\begin{table}[h]
  \centering
    \caption{Evaluation of the HDM. Accuracy is reported at all three hierarchical levels of EUNIS habitat classification (level 1 being the broader and level 3 the finer). Retaining more species from the deep-SDM predictions slightly improves classification performance across all levels.}
  \begin{tabular}{@{}l@{\hspace{1cm}}ccc@{}}
    \toprule
    Top-SDM predictions & Level 1 & Level 2 & Level 3 \\
    \midrule
    First 50 species & 75.05\% & 61.29\% & 42.78\% \\
    First 100 species & \textbf{76.30\%} & \textbf{62.68\%} & \textbf{44.72\%} \\
    \bottomrule
  \end{tabular}
  \label{tab:hdm evaluation}
\end{table}

Experiments have shown that Pl@ntBERT, through its ability to model complex inter-species relationships, is able to outperform expert systems ~\cite{chytry2020eunis} (+5.54\%) and tabular deep learning ~\cite{leblanc2024deep} (+1.14\%) methods. Overall, the measured accuracy was 76\% at level 1 of the EUNIS classification (8 broad habitat groups covered), 63\% at level 2 (34 habitat groups covered), and 45\% at level 3 (200 habitat types covered). In Table \ref{tab:hdm evaluation}, we report the full performance evaluation with respect to the number of species that has been kept from the SDM predictions. 

The model benefits from the fact that the SDM provides richer information, i.e., a calibrated softmax. Those probabilities are used directly as input in Pl@ntBERT, with predicted species being ordered in descending probability order in each sentence. This is a “reciprocal rank encoding method” but uses the probability score as the ranking function instead of the spatial coverage.

%% file: sec/4_conclusion.tex
\section{Conclusion}
\label{sec:conclusion}

This work presents a multi-modal deep learning framework based on species distribution modeling (SDM), biodiversity indicators calculation, and habitat classification for high-resolution biodiversity mapping across Europe. Using remote sensing, climate variables, and species occurrence data, we provide a comprehensive, fine-scale view on species distributions, ecosystem diversity, and habitat types at an unprecedented 50$\times$50m resolution at this scale. Our approach enables previously infeasible large-scale ecological assessments, offering new tools for biodiversity monitoring, conservation planning, and land-use management.

\textbf{The Species Distribution Maps}, generated using deep-SDM, effectively predict species occurrences by combining satellite imagery, climate time-series, and species records from GBIF and EVA. These maps provide baseline data for over 5,5k plant species, supporting efforts to track species distributions, monitor ecological shifts, and guide conservation policies.

\textbf{The Biodiversity Indicator Maps} provide insights into species richness, the presence of endangered or invasive species, as well as other key ecological metrics. These maps help identify biodiversity hotspots, vulnerable ecosystems, and priority areas for conservation.

\textbf{The Habitat Maps}, created by coupling SDM predictions with Pl@ntBERT, classify EUNIS habitat types across Europe. While these maps enhance the understanding of ecosystem distributions and habitat changes, challenges remain in classifying habitats at EUNIS Level 3, partly due to inconsistencies in expert-labeled training data.

Despite large contributions, several limitations remain. The reliance on species occurrence data from citizen science platforms and surveys introduces spatial biases, as certain regions and species are better documented than others (e.g, PO data are biased toward appealing species and PA data have limited geographic coverage). Additionally, prediction uncertainties persist, particularly in areas with low observation density or rapidly changing environmental conditions. The classification of habitats is further constrained by potential inconsistencies in EUNIS labeling, impacting the reliability of fine-scale habitat predictions. Finally, the multimodal nature and the size of the dataset require considerable computational resources for model training.

%% file: sec/X_acknowledgments.tex
\section*{Acknowledgments}
\label{sec:acknowledgments}

The research described in this paper was funded by the European Commission through the GUARDEN (safeGUARDing biodivErsity aNd critical ecosystem services across sectors and scales) and MAMBO (Modern Approaches to the Monitoring of BiOdiversity) projects. These projects received funding from the European Union’s Horizon Europe research and innovation programme under grant agreements 101060693 (start date: 01/11/2022; end date: 31/10/2025) and 101060639 (start date: 01/09/2022; end date: 31/08/2026), respectively. Further models developed based on this methodology will directly meet the needs of the European biodiversity strategy for 2030 through those projects. They will be used in particular to enhance the biodiversity maps at the European scale. The content of this paper reflects the views only of the authors, and the European Commission cannot be held responsible for any use which may be made of the information contained therein. 

The authors are grateful to the OPAL infrastructure from Université Côte d’Azur for providing resources and support. This work was granted access to the high-performance computing resources of IDRIS (Institut du Développement et des Ressources en Informatique Scientifique) under the allocation 2023-AD010113641R1 made by GENCI (Grand Equipement National de Calcul Intensif). 

Our major thanks go to thousands of European vegetation scientists of several generations who collected the original vegetation-plot data in the field and made their data available to others and those who spent myriad hours digitizing data and managing the databases in the EVA. Vegetation plots data for this study were provided by Sylvain Abdulhak, Alicia Acosta, Emiliano Agrillo, Pierangela Angelini, Iva Apostolova, Olivier Argagnon, Fabio Attorre, Svetlana Aćić, Christian Berg, Ariel Bergamini, Erwin Bergmeier, Idoia Biurrun, Maxim Bobrovsky, Steffen Boch, Gianmaria Bonari, Anne Bonis, Zoltán Botta-Dukát, Jan-Bernard Bouzillé, Helge Bruelheide, Vanessa Bruzzaniti, Juan Antonio Campos, Andraž Čarni, Maria Laura Carranza, Laura Casella, Alessandro Chiarucci, Andrei Chuvashov, Milan Chytrý, János Csiky, Mirjana Krstivojević Ćuk, Renata Ćušterevska, Olga Demina, Jürgen Dengler, Panayotis Dimopoulos, Dmytro Dubyna, Tetiana Dziuba, Alexei Egorov, Rasmus Ejrnæs, Franz Essl, Jörg Ewald, Giuliano Fanelli, Federico Fernández-González, Úna FitzPatrick, Xavier Font, Gianpietro Giusso del Galdo, Emmanuel Garbolino, Itziar García-Mijangos, Rosario G Gavilán, Jean-Michel Genis, Michael Glaser, Valentin Golub, Friedemann Goral, Jean-Claude Gégout, Behlül Güler, Rense Haveman, Stephan Hennekens, Adrian Indreica, Maike Isermann, Ute Jandt, Jan Jansen, Florian Jansen, John Janssen, Anni Kanerva Jašková, Borja Jiménez-Alfaro, Martin Jiroušek, Veronika Kalníková, Ali Kavgacı, Larisa Khanina, Ilona Knollová, Vitaliy Kolomiychuk, Łukasz Kozub, Daniel Krstonošić, Helmut Kudrnovsky, Anna Kuzemko, Filip Küzmič, Zygmunt Kącki, Flavia Landucci, Igor Lavrinenko, Jonathan Lenoir, Armin Macanović, Corrado Marcenò, Aleksander Marinšek, Marco Massimi, Ruth Mitchell, Jesper Erenskjold Moeslund, Pavel Novák, Vladimir Onipchenko, Robin Pakeman, Hristo Pedashenko, Tomáš Peterka, Remigiusz Pielech, Vadim Prokhorov, Ricarda Pätsch, Aaron Pérez-Haase, Valerius Rašomavičius, Maria Pilar Rodríguez-Rojo, John Rodwell, Iris de Ronde, Eszter Ruprecht, Solvita Rūsiņa, Michele De Sanctis, Joop Schaminée, Joachim Schrautzer, Ingrid Seynave, Pavel Shirokikh, Jozef Šibík, Urban Šilc, Željko Škvorc, Desislava Sopotlieva, Angela Stanisci, Milica Stanišić-Vujačić, Zora Dajić Stevanović, Danijela Stešević, Jens-Christian Svenning, Grzegorz Swacha, Irina Tatarenko, Ioannis Tsiripidis, Ruslan Tsvirko, Pavel Dan Turtureanu, Domas Uogintas, Emin Uğurlu, Milan Valachovič, Kiril Vassilev, Roberto Venanzoni, Sophie Vermeersch, Risto Virtanen, Denys Vynokurov, Lynda Weekes, Wolfgang Willner, Thomas Wohlgemuth, Svitlana Yemelianova, and Dominik Zukal.